\documentclass[10pt,twocolumn,letterpaper]{article}

\usepackage{wacv}              % To produce the CAMERA-READY version
\usepackage{graphicx}
\usepackage{amsmath}
\usepackage{amssymb}
\usepackage{booktabs}
\usepackage{multirow}

\usepackage[pagebackref,breaklinks,colorlinks]{hyperref}

\usepackage[capitalize]{cleveref}
\crefname{section}{Sec.}{Secs.}
\Crefname{section}{Section}{Sections}
\Crefname{table}{Table}{Tables}
\crefname{table}{Tab.}{Tabs.}

\def\wacvPaperID{997} % *** Enter the WACV Paper ID here
\def\confName{WACV}
\def\confYear{2025}

\begin{document}

%%%%%%%%% TITLE - PLEASE UPDATE
\title{HSDA: High-frequency Shuffle Data Augmentation for Bird's-Eye-View Map Segmentation}

\author{Calvin Glisson\\
School of Computer Science and Engineering\\
California State University, San Bernardino\\
San Bernardino, USA \\
{\tt\small cglis001@ucr.edu}
% For a paper whose authors are all at the same institution,
% omit the following lines up until the closing ``}''.
% Additional authors and addresses can be added with ``\and'',
% just like the second author.
% To save space, use either the email address or home page, not both
\and
Qiuxiao Chen\\
School of Computer Science and Engineering\\
California State University, San Bernardino\\
San Bernardino, USA \\
{\tt\small Qiuxiao.Chen@csusb.edu}
}
\maketitle

%%%%%%%%% ABSTRACT
\begin{abstract}
Autonomous driving has garnered significant attention in recent research, and Bird's-Eye-View (BEV) map segmentation plays a vital role in the field, providing the basis for safe and reliable operation. While data augmentation is a commonly used technique for improving BEV map segmentation networks, existing approaches predominantly focus on manipulating spatial domain representations. In this work, we investigate the potential of frequency domain data augmentation for camera-based BEV map segmentation. We observe that high-frequency information in camera images is particularly crucial for accurate segmentation. Based on this insight, we propose High-frequency Shuffle Data Augmentation (HSDA), a novel data augmentation strategy that enhances a network's ability to interpret high-frequency image content. This approach encourages the network to distinguish relevant high-frequency information from noise, leading to improved segmentation results for small and intricate image regions, as well as sharper edge and detail perception. Evaluated on the nuScenes dataset, our method demonstrates broad applicability across various BEV map segmentation networks, achieving a new state-of-the-art mean Intersection over Union (mIoU) of 61.3\% for camera-only systems. This significant improvement underscores the potential of frequency domain data augmentation for advancing the field of autonomous driving perception. Code has been released: \url{https://github.com/Zarhult/HSDA}
\end{abstract}

%%%%%%%%% BODY TEXT
\section{Introduction}
\label{sec:intro}

Bird's-Eye-View (BEV) map segmentation processes sensor data to generate a top-down semantic map of a vehicle's surroundings, classifying grid cells into categories such as drivable areas, pedestrian crossings, and walkways. BEV map segmentation has garnered substantial research interest due to its pivotal role in applications that include autonomous driving, robotics, and autonomous warehouse navigation. Specifically, BEV semantic maps provide foundational input for critical tasks such as motion prediction \cite{wu2020motionnet, hu2021fiery, fang2023tbp}, trajectory planning \cite{hu2023planning}, decision making \cite{liu2023bird}, and control learning \cite{chitta2021neat} in autonomous systems.

The paramount importance of BEV segmentation for safe and efficient operation has prompted extensive research to enhance its performance, accuracy, and robustness \cite{chen2024residual,liu2023bevfusion,li2022bevformer,chen2023bevseg,philion2020lift,zhou2022cross,ge2023metabev,ji2023ddp}. Early work \cite{philion2020lift} introduced an end-to-end approach using depth estimation and voxel-based techniques. Recent studies have expanded on this by exploring transformers \cite{li2022bevformer,zhou2022cross}, denoising diffusion models \cite{ji2023ddp}, and multi-modal feature fusion \cite{zhang2022beverse,xie2022m} to advance spatial domain capabilities.

\begin{figure*}
    \centering
    \includegraphics[width=0.95\linewidth]{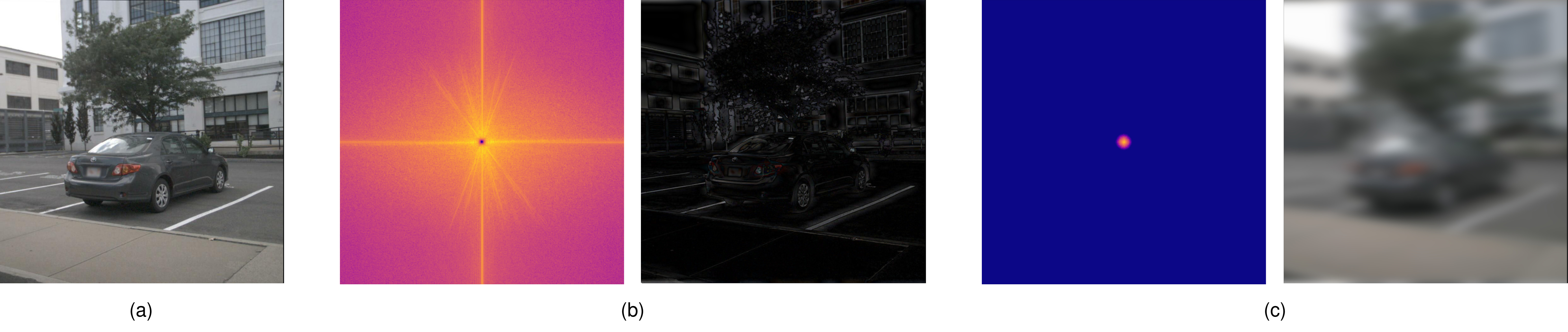}
    \captionsetup{width=\textwidth}
    \caption{Illustration of high-frequency spectrum, low-frequency spectrum and the corresponding images. (a) Original image. (b) High-frequency spectrum and corresponding image. The image becomes primarily dark but retains key edges and outlines.  (c) Low-frequency spectrum and corresponding image. All sharp edges and rapid visual changes are removed, effectively blurring the image.}
    \label{fig:hf_lf}
\end{figure*}

\begin{table*}
  \begin{center}
    {\normalsize
\begin{tabular}{lrrrrrrr}
\toprule
Input Type & drivable\_area & ped\_crossing & walkway & stop\_line & carpark\_area & divider & mean\\
\midrule
Original & 81.2 & 54.6 & 58.9 & 48.5 & 52.1 & 51.9 & 57.9\\
Low-Frequency Only & 70.1 & 37.0 & 43.2 & 31.9 & 36.2 & 36.6 & 42.5\\ 
High-Frequency Only & 79.0 & 50.8 & 55.3 & 44.3 & 48.6 & 48.9 & 54.5\\
\bottomrule
\end{tabular}
}
\end{center}
\caption{Analysis of the impact of low-frequency and high-frequency information on the baseline model.}
\label{tab:freq_comparison}
\end{table*}

This paper explores the underutilized potential of the frequency domain to enhance information extraction from input data. Specifically, it examines the complementary roles of low-frequency and high-frequency components in image representation. Low-frequency components capture gradual changes and are concentrated at the spectrum's center, while high-frequency components highlight edges, textures, and fine details, essential for tasks like object detection and segmentation, such as identifying stop lines and pedestrian crossings. These components are illustrated in Figure \ref{fig:hf_lf}.

To initially assess the role of low-frequency and high-frequency information in autonomous driving, we modified the BEV map segmentation baseline model to use only one frequency component during training and inference. As shown in Table \ref{tab:freq_comparison}, models limited to low-frequency data suffered significant accuracy loss, while those using high-frequency data showed only a minor decline. This is because high-frequency details, such as edges and textures, are crucial for defining region boundaries, whereas low-frequency data mostly contain smooth regions and lack clear dividing lines. These findings indicate that BEV map segmentation primarily depends on the high-frequency component of camera image information.

We therefore propose a High-frequency Shuffle Data Augmentation (HSDA) method for multi-view BEV map segmentation, utilizing the Fast Fourier Transform (FFT) and Gaussian filters to separate an image into high and low-frequency spectra. We then augment the high-frequency component by randomly shuffling the dominant high frequencies to introduce controlled noise, while keeping the original BEV map. Training on both original and augmented data helps the model learn the correlation between high-frequency elements and the BEV map, improving segmentation performance by focusing on essential high-frequency components.

Our data augmentation technique offers several key advantages. Notably, it requires no modifications to the baseline network architecture or additional parameters to achieve substantial improvement. Our contributions are as follows: 1) We first assert the significance of frequency information in BEV map segmentation. 2) We propose an effective and widely applicable data augmentation method, High-frequency Shuffle Data Augmentation (HSDA). 3) The HSDA method achieves state-of-the-art performance on the nuScenes map segmentation benchmark, surpassing previous approaches by at least 1.6\% mIoU.

%-------------------------------------------------------------------------

\begin{figure*}
    \centering
    \includegraphics[width=0.9\linewidth]{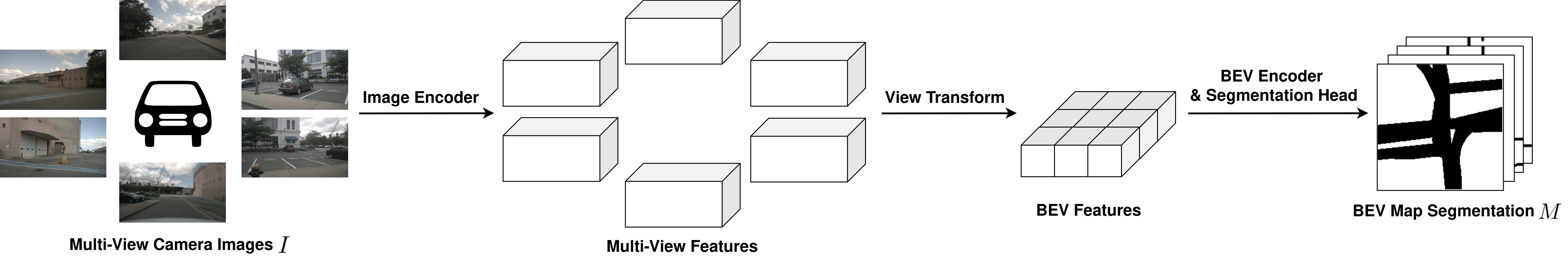}
    \caption{Overview of our baseline network architecture. It begins by processing multi-view camera images using an image encoder to extract features. These features are then transformed into the BEV space using a view transformation module that leverages camera intrinsics and extrinsics. Subsequently, a BEV encoder processes the transformed features, which are then passed to a segmentation head to generate the final BEV map segmentation predictions.}
    \label{fig:outline}
\end{figure*}

\section{Related Work}
\label{sec:related}
\subsection{BEV Map Segmentation}
BEV map segmentation is primarily done by integrating information across multiple camera images that provide a view of the surroundings. Traditionally, segmentation is applied directly to images \cite{borse2021hs3, borse2022panoptic, borse2021inverseform, hu2022learning, zhang2022auxadapt, zhang2022perceptual, dong2020real}. Subsequent work has used homography transformation to convert from the camera image view to BEV \cite{garnett20193d, loukkal2021driving, zhu2021monocular, chen2022monocular}. However, homography transformation introduces substantial error, motivating alternative approaches. Frequently, these approaches perform the conversion to BEV as an end-to-end learning task. For instance, LSS\cite{philion2020lift} does so by predicting depth distributions for each pixel.

On this basis, various new ideas have been explored. BEVSeg \cite{chen2023bevseg} introduces a low-complexity attention-based method for weighing the importance of spatial features. MetaBEV\cite{ge2023metabev} increases network robustness to sensor corruption or failure, and alleviates task conflict with the proposed $M^{2}oE$ structure. BEVFormer\cite{li2022bevformer} proposes a transformer-based network that applies attention to spatial and temporal information used for 3D object detection and BEV map segmentation. PETRv2\cite{liu2023petrv2} introduces task-specific queries to support tasks including BEV segmentation and 3D detection while utilizing temporal information from previous frames. DDP \cite{ji2023ddp} explores perception through denoising diffusion, offering dynamic inference, inference trajectory and uncertainty awareness. X-Align \cite{borse2023x} enhances feature fusion and alignment between joint LiDAR and camera modalities. 
Furthermore, a novel residual graph convolutional module \cite{chen2024residual} has been applied to segmentation, helping the model estimate contextual relationships between regions of the global features. 

Previous works have explored various methods to refine BEV-based networks, but Fourier transform applications to camera images used for BEV feature generation remain unexplored. Our work highlights the value of the Fourier transform and frequency domain in map segmentation.

\subsection{Data Augmentation in Autonomous Driving}

Data augmentation is a popular research topic in autonomous driving, and has been applied to sensor data, such as LiDAR point clouds and images. In the context of LiDAR point clouds, RS-Aug \cite{an2023rs} uses realistic simulations to leverage unlabeled LiDAR data. Pattern-aware ground truth sampling \cite{hu2021pattern} remedies the relative lack of LiDAR data for objects far from the ego vehicle. Regarding image-based perception, both monocular and multi-view approaches have adopted data augmentation techniques. For instance, 2D data augmentation techniques including random translation and resizing have been adapted to the monocular 3D object detection task\cite{jia2024enhancing}. With multi-view camera images as input, BEVDet\cite{huang2021bevdet}, BEVDepth\cite{li2023bevdepth}, and MetaBEV\cite{ge2023metabev} apply data augmentation strategies such as random flipping and random scaling to input images. BEVDet also introduces data augmentation methodologies applied to the BEV features to combat overfitting. However, these data augmentation methods are applied only within the spatial domain, neglecting the frequency domain.

\subsection{Frequency Domain Data Augmentation}

Frequency domain data augmentation has proven beneficial in various deep learning domains. Notably, frequency warping has been utilized in speech recognition as a form of vocal tract length perturbation \cite{iglesias2023data}.  Additionally, frequency domain-based strategies have been employed in time series representation learning, such as in TimesURL \cite{liu2024timesurl} and Dominant Shuffle \cite{zhao2024dominant}. These methods leverage Fourier transforms to augment data in the frequency domain before converting back to the original domain. 
Specifically, Dominant Shuffle perturbs the top $K$ frequencies by magnitude, while TimesURL employs a self-supervised approach that incorporates the construction of double Universums and data reconstruction. Furthermore, FDA \cite{he2024frequency} has applied frequency domain augmentation to images in Vision-and-Language navigation. However, to our knowledge, our work represents the first exploration of frequency domain data augmentation in the field of autonomous driving.

%-------------------------------------------------------------------------

\section{Proposed HSDA}
\label{sec:method}

\subsection{Problem Formulation and Baseline Overview}
\begin{figure*}
    \centering
    \includegraphics[width=0.85\linewidth]{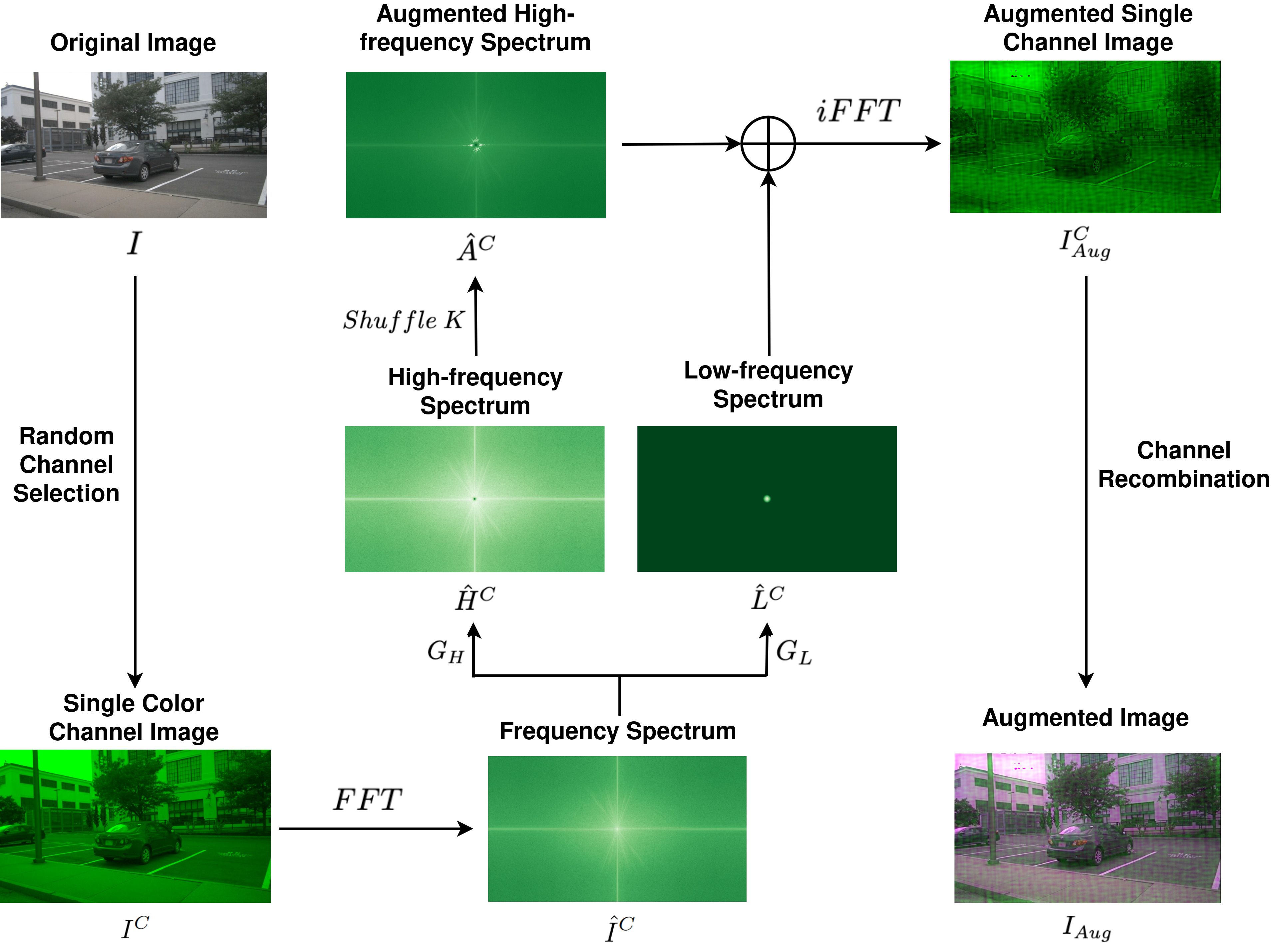}
    \captionsetup{width=\textwidth}
    \caption{The proposed High-frequency Shuffle Data Augmentation (HSDA) method introduces perturbations in the high-frequency domain. HSDA operates on a randomly selected color channel, applying the Fast Fourier Transform (FFT) and filtering to obtain high-frequency and low-frequency components. The most salient $K$ frequencies within the high-frequency spectrum are shuffled to introduce controlled noise, which we emphasize in $\hat{A}^C$ for ease of visualization.
    Recombining with the original low-frequency spectrum and applying the inverse Fast Fourier Transform (iFFT) yields the augmented single-channel image. This replaces the original channel to generate the final augmented image. In this example, the green channel is randomly chosen from RGB channels for shuffling, causing the green color information in the final image to be perturbed. This creates a grid-like pattern of regions with excess or insufficient green intensity.}
    \label{fig:HSDA}
\end{figure*}
Bird's-eye view map segmentation aims to construct a model that takes multi-view camera images $I$ as input and produces corresponding BEV segmentation maps $M$ of the ego vehicle's surroundings. Formally, the input $I$ is defined as a set of $N_{view}$ camera views, i.e., $I=\{I_i\}_{i=1}^{N_{view}}$, where each $I_i \in \mathbb{R}^{NC \times H \times W}$, represents an individual image with $NC$ color channels, height $H$, and width $W$. The output $M \in \mathbb{R}^{SC \times H_{pred} \times W_{pred}}$ is a segmentation map with $SC$ semantic classes, height $H_{pred}$, and width $W_{pred}$, providing a perspective of the environment's semantic layout.

In general, the HSDA method is applicable to a wide range of BEV models, including the streamlined baseline, based on BEVDet \cite{huang2021bevdet}. As shown in Figure \ref{fig:outline}, the baseline architecture consists of four primary modules: an image encoder, a view transform, a BEV encoder, and a segmentation head. With the exception of the initial three components from BEVDet, we substitute BEVDet's final component, namely the 3D object detection head, with our custom-designed map segmentation head. This modification results in a streamlined, modular map segmentation architecture.

\subsection{High-frequency Shuffle Data Augmentation}
\label{ssec:augmentation}

As shown in Figure \ref{fig:HSDA}, we use a single image $I$, selected from the set of multi-view images, to exemplify our approach. Our proposed data augmentation method, HSDA, introduces random perturbations into the dominant frequencies within the high-frequency component of the single image. To accomplish this, we first randomly select a color channel $C\in \{0, 1, 2\}$ (corresponding to red, green, and blue respectively) from image $I$, denoted as $I^C$. Subsequently, we transform $I^C$ into the frequency domain via the Fast Fourier Transform (FFT):
\begin{equation}FFT(I^C) = \hat{I}^C\end{equation} 
We then decompose the frequency spectrum $\hat{I}^C$ into its low-frequency and high-frequency components, denoted as $\hat{L}^C$ and $\hat{H}^C$ respectively, using Gaussian low-pass and high-pass filters $G_L$ and $G_H$:
\begin{equation}\hat{L}^C = G_L\odot \hat{I}^C\end{equation} 
\begin{equation}\hat{H}^C = G_H\odot \hat{I}^C\end{equation} where $\odot$ denotes the element-wise Hadamard product. Our low-pass Gaussian filter is a standard Gaussian function of the form \begin{equation}{G_L =  e^{-\frac{x^2+y^2}{2D^2}}}\label{eq:g_l}\end{equation} in the frequency domain, where $x$ and $y$ are pixel coordinates relative to the central origin. The $D$ parameter is manually chosen, and determines the threshold for delineation of high-frequency and low-frequency components. The high-pass filter is simply $G_H=1-G_L$. The ablation study of the D value is shown in Sec \ref{sssec:ablation}.

\begin{table*}
\normalsize
\centering
\begin{tabular}{lrrrrrrr}
\toprule
$K$ & drivable\_area & ped\_crossing & walkway & stop\_line & carpark\_area & divider & mean\\
\midrule
1000 & 81.0 & 56.4 & 59.6 & 51.8 & 52.0 & 53.5 & 59.0\\
2000 & 81.2 & \textbf{56.5} & \textbf{59.7} & \textbf{52.1} & \textbf{53.2} & \textbf{54.3} & \textbf{59.5}\\
3000 & \textbf{81.3} & 56.0 & 59.6 & 51.6 & 51.3 & 53.7 & 58.9 \\
\bottomrule
\end{tabular}

\caption{Comparison of $K$ values. All results are applications of HSDA to the baseline network with the specified $K$ value used for image augmentation.}
\label{tab:top_k}
\end{table*}
\begin{table*}
\normalsize
\centering
\begin{tabular}{lrrrrrrr}
\toprule
$D$ & drivable\_area & ped\_crossing & walkway & stop\_line & carpark\_area & divider & mean\\
\midrule
5 & 80.7 & 55.8 & 59.1 & 51.1 & 51.6 & 53.6 & 58.7\\
10 & 81.2 & \textbf{56.5} & 59.7 & \textbf{52.1} & \textbf{53.2} & \textbf{54.3} & \textbf{59.5}\\
15 & \textbf{81.4} & \textbf{56.5} & \textbf{59.8} & 51.8 & 52.7 & 54.1 & 59.4 \\
\bottomrule
\end{tabular}

\caption{Comparison of $D$ values. All results are applications of HSDA to the baseline network with the specified $D$ value used for image augmentation.}
\label{tab:best_d}
\end{table*}
Following the separation of the low and high-frequency components, the top $K$ frequencies in the high-frequency spectrum are identified based on their magnitude and shuffled to create an augmented high-frequency spectrum $\hat{A}^C$:
\begin{equation}Shuffle(\hat{H}^C,K)=\hat{A}^C\end{equation}
More precisely, the $Shuffle$ operation executes a randomized swap for each pixel within the pool of the top $K$ frequencies, where $K$ represents a predetermined value. The ablation study of the $K$ value is shown in Sec \ref{sssec:ablation}.
Finally, the augmented high-frequency component $\hat{A}^C$ is combined with the original low-frequency component $\hat{L}^C$, and an inverse Fast Fourier Transform (iFFT) is applied to obtain the augmented channel $I^C_{Aug}$ in the spatial domain:
\begin{equation}iFFT(\hat{L}^C+\hat{A}^C)=I^C_{Aug}\end{equation}
This augmented channel replaces the original, resulting in the final augmented image, $I_{Aug}$, which shares the same map segmentation ground truth as the input image $I$. Figure \ref{fig:HSDA} illustrates the augmentation process.

As illustrated in Figure \ref{fig:HSDA}, the augmentation process tends to generate grid-like artifacts within the image.  The shuffling operation introduces a misalignment within the randomly selected color channel, contrasting with the other channels that remain unaltered during training. This encourages the network to learn the relationship between the high-frequency shuffled information and the intact image data. From a holistic perspective, the displacement of color information from its original location results in most image regions exhibiting either an excess or a deficiency of the chosen color, thus creating the green tinting effect observed in $I_{Aug}$ in Figure \ref{fig:HSDA}.

%-------------------------------------------------------------------------

\section{Experimental Results}
\label{sec:experimental}

\subsection{Experiment Setup}
\label{ssec:setup}

\textbf{Datasets:} We evaluate our network's performance using the large-scale autonomous driving dataset nuScenes \cite{caesar2020nuscenes}. With data collected from 1,000 scenes in diverse cities, the full dataset contains 1,400,00 camera images and comprehensive map information for all scenes. 700 of the scenes belong to the training set, with the remaining 300 split evenly between the validation and testing sets. 
The evaluation of all our models was conducted exclusively on the validation set, as the leaderboard, which would provide access to test results, does not currently support the map segmentation task.
Each data sample provides six camera images that yield a comprehensive view of the vehicle's surroundings, accompanied by semantic map annotations providing the ground truth semantic map segmentation for 11 classes. To maintain consistency with recent state-of-the-art research \cite{liu2023bevfusion}\cite{borse2023x}\cite{ge2023metabev}\cite{ji2023ddp}\cite{chen2024residual}, we focus on maximizing the IoU of our predictions for six key semantic classes: drivable area, pedestrian crossing, walkway, stop line, carpark area, and divider.

\textbf{Data Augmentation:}
We apply the data augmentation techniques employed in BEVDet \cite{huang2021bevdet}. This includes random flipping, rotation, scaling, and cropping of the image data, as well as random flipping, rotation, and scaling of BEV features.
In addition to these established methods, we introduce our proposed HSDA augmentation, as detailed in Sec \ref{ssec:augmentation}. Prior to training, we apply HSDA to all nuScenes camera images and combine these augmented images with the original data to form the final training dataset.

\begin{table*}
  \begin{center}
    {\normalsize{
\begin{tabular}{lrrrrrrr}
\toprule
& drivable\_area & ped\_crossing & walkway & stop\_line & carpark\_area & divider & mean\\
\midrule
BEVFusion & 81.7 & 54.8 & 58.4 & 47.4 & 50.7 & 46.4 & 56.6 \\
BEVFusion + HSDA & \textbf{82.2} & \textbf{57.7} & \textbf{60.0} & \textbf{51.3} & \textbf{53.5} & \textbf{47.9} & \textbf{58.8} \\
\midrule
Baseline & \textbf{81.2} & 54.6 & 58.9 & 48.5 & 52.1 & 51.9 & 57.9\\
Baseline + HSDA & \textbf{81.2} & \textbf{56.5} & \textbf{59.7} & \textbf{52.1} & \textbf{53.2} & \textbf{54.3} & \textbf{59.5} \\ \midrule
RGC & 81.7 & 57.1 & 60.5 & 51.7 & 53.8 & 53.5 & 59.7 \\
RGC + HSDA & \textbf{82.3} & \textbf{58.3} & \textbf{61.5} & \textbf{54.8} & \textbf{55.5} & \textbf{55.3} & \textbf{61.3} \\
\bottomrule
\end{tabular}
}}
\end{center}
\caption{Ablation study of BEV map segmentation models before and after applying HSDA.}
\label{tab:robust}
\end{table*}

\begin{table*}
  \begin{center}
    {\normalsize{
\begin{tabular}{lrrrrrrr}
\toprule
& drivable\_area & ped\_crossing & walkway & stop\_line & carpark\_area & divider & mean\\
\midrule
Baseline & \textbf{81.2} & 54.6 & 58.9 & 48.5 & 52.1 & 51.9 & 57.9\\
Baseline + FDA & 81.0 & 55.5 & \textbf{59.7} & 51.6 & 51.1 & 53.4 & 58.7\\
Baseline + HSDA & \textbf{81.2} & \textbf{56.5} & \textbf{59.7} & \textbf{52.1} & \textbf{53.2} & \textbf{54.3} & \textbf{59.5}\\
\midrule
RGC & 81.7 & 57.1 & 60.5 & 51.7 & 53.8 & 53.5 & 59.7 \\
RGC + FDA & 81.7 & 57.4 & 60.8 & 54.1 & 53.2 & 55.0 & 60.4\\
RGC + HSDA & \textbf{82.3} & \textbf{58.3} & \textbf{61.5} & \textbf{54.8} & \textbf{55.5} & \textbf{55.3} & \textbf{61.3} \\
\bottomrule
\end{tabular}
}}
\end{center}
\caption{Comparison of perturbation methods for data augmentation.}
\label{tab:other_DA}
\end{table*}

\textbf{Implementation and Training:}
To optimize computational efficiency, all input images are downsampled to a resolution of $256\times704$ before network processing. All of our models are trained for 20 epochs using CBGS\cite{zhu2019class}. Optimization is performed with the AdamW optimizer \cite{loshchilov2017decoupled} with a learning rate of 2e-4 and a cyclic learning rate policy \cite{smith2017cyclical}.

\subsection{Quantitative Results}
\label{ssec:quantitative}

\subsubsection{Ablation Study: }

\label{sssec:ablation}
\hspace*{1em}\textbf{Values of $\mathbf{K}$: } A critical factor in the effective implementation of HSDA is the selection of an appropriate value for $K$, representing the number of shuffled pixels in the frequency spectrum. While a small $K$ value may result in overly subtle perturbations, a large $K$ value risks distorting the image too severely. To identify the optimal $K$ value for our network, we trained three variants of the Baseline+HSDA network with $K$ values of 1000, 2000, and 3000. The results, presented in Table \ref{tab:top_k}, indicate that a $K$ value of 2000 strikes the best balance. Therefore, we adopt this value for all of our subsequent experiments.

\textbf{Values of $\mathbf{D}$: }
The separation of low and high frequencies is at the core of our method, and it is therefore necessary to choose an effective value of $D$ in equation \eqref{eq:g_l} for our Gaussian filters. A higher value of $D$ corresponds to a low-pass filter with a higher cutoff frequency, attenuating the high-frequency spectrum, while a lower value does the opposite. 
Table \ref{tab:best_d} presents the results obtained with varying $D$ values. We observe that a $D$ value of 5 notably diminishes HSDA's effectiveness, even proving detrimental for certain classes. While $D$ values of 10 and 15 produce comparable results, 10 emerges as the superior choice overall. Due to this result, we opt for a $D$ value of 10 for all of our subsequent experiments.

\textbf{Generalizable to different models: }
As illustrated in Table \ref{tab:robust}, the application of HSDA consistently improves performance across different network architectures. Specifically, BEVFusion, our Baseline model, and RGC demonstrate mIoU gains of 2.2\%, 1.6\%, and 1.6\% respectively when HSDA is applied. In our experiments, the application of HSDA improves performance across nearly all classes regardless of the model it is applied to. Moreover, there are no instances where the segmentation accuracy of a class is diminished by HSDA.

\textbf{Comparison with Existing Data Augmentation:}
Table \ref{tab:other_DA} presents a comparison with Frequency-enhanced Data Augmentation (FDA) \cite{he2024frequency}, previously explored in the Vision-and-Language navigation task. In contrast to FDA, which perturbs the high-frequency component by substituting it with that of another training image, our HSDA method randomly shuffles the dominant high frequencies within the same image. Furthermore, HSDA achieves a 1.6\% improvement over both the baseline and RGC networks, surpassing the 0.8\% and 0.7\% improvements offered by FDA.
Additionally, HSDA demonstrates superior performance across all individual classes compared to FDA, with the exception of the "walkway" category on the baseline network where the methods achieve parity. Our method offers both enhanced performance and greater ease of implementation, as it can be applied independently to each input image without the need for additional interference images.

\subsubsection{State-Of-The-Art Method Comparison: }
\label{sssec:SOTA}
\begin{table*}[h!]
\normalsize
\centering
\begin{tabular}{lrrrrrrr}
\hline
            & drivable\_area & ped\_crossing & walkway & stop\_line & carpark\_area & divider & mean \\ \hline
OFT \cite{roddick2018orthographic}        & 74.0            & 35.3          & 45.9    & 27.5       & 35.9          & 33.9    & 42.1 \\
LSS \cite{philion2020lift}         & 75.4          & 38.8          & 46.3    & 30.3       & 39.1          & 36.5    & 44.4 \\
CVT \cite{zhou2022cross}        & 74.3          & 36.8          & 39.9    & 25.8       & 35.0            & 29.4    & 40.2 \\
BEVFusion \cite{liu2023bevfusion}   & 81.7          & 54.8          & 58.4    & 47.4       & 50.7          & 46.4    & 56.6 \\ 
X-Align \cite{borse2023x} & 82.4 & 55.6 & 59.3 & 49.6 & \textit{53.8} & 47.4 & 58.0 \\
MetaBEV \cite{ge2023metabev}    & \textit{83.3} & 56.7 & 61.4 & 50.8 & \textbf{55.5} & 48.0 & 59.3 \\
DDP (step 3) \cite{ji2023ddp} & \textbf{83.6} & \textbf{58.3} & \textbf{61.8} & \textit{52.3} & 51.4 & 49.2 & 59.4 \\
RGC \cite{chen2024residual}  & 81.7 & \textit{57.1} & 60.5 & 51.7 & \textit{53.8} & \textit{53.5} & \textit{59.7} \\ \hline
Ours (RGC+HSDA)  & 82.3 & \textbf{58.3} & \textit{61.5} & \textbf{54.8} & \textbf{55.5} & \textbf{55.3} & \textbf{61.3} \\ \hline
\end{tabular}
\caption{BEV map segmentation SOTA model comparison. Bold font represents the best performance. Italics represent the second best performance.}
\label{tab:SOTA}
\end{table*}
\begin{table}[h!]
\normalsize
\centering
\begin{tabular}{lrrr}
\hline
            & drivable\_area & divider & mean \\ \hline
BEVFormer \cite{li2022bevformer}   & 80.7 & 21.3 & 51.0\\ 
PETRv2 \cite{liu2023petrv2} & \textbf{83.3} & 44.8 & 64.1 \\ \hline
Ours (RGC+HSDA)  & 81.5 & \textbf{52.3} & \textbf{66.9} \\ \hline
\end{tabular}
\caption{BEV map segmentation SOTA model comparison restricted to drivable area and divider. Bold font represents the best performance.}
\label{tab:SOTA_TwoClass}
\end{table}
\begin{table*}[]
\normalsize
\centering
\begin{tabular}{c|ccc|ccc|ccc|ccc}
\hline
\multirow{2}{*}{} & \multicolumn{3}{c|}{Pedestrian $AP_{3D|IoU\geq 0.5}$} & \multicolumn{3}{c|}{Cyclist $AP_{3D|IoU\geq 0.5}$} & \multicolumn{3}{c|}{Car $AP_{3D|IoU\geq 0.7}$} & \multicolumn{3}{c}{Mean} \\
& Easy & Mod. & Hard & Easy & Mod. & Hard & Easy & Mod. & Hard & Easy & Mod. & Hard \\ \hline
MonoCon & 3.82 & 3.21 & 2.58 & \textbf{6.44} & \textbf{3.54} & \textbf{3.05} & 22.88 & 16.60 & 14.58 & 11.05 & 7.78 & 6.74 \\
MonoCon+HSDA & \textbf{9.01} & \textbf{6.76} & \textbf{5.36} & 5.93 & 3.18 & 2.97 & \textbf{24.38} & \textbf{17.25} & \textbf{15.10} & \textbf{13.11} & \textbf{9.06} & \textbf{7.81} \\ \hline
\end{tabular}
\caption{Results of applying HSDA to the monocular 3D object detection task with MonoCon.}
\label{tab:monocon}
\end{table*}
Tables \ref{tab:SOTA} and \ref{tab:SOTA_TwoClass} present our results and compare them with ten recent SOTA networks. To fully exploit the capabilities of HSDA, we apply it to the previous state-of-the-art (SOTA) camera-only BEV map segmentation model, RGC, as our proposed method. To ensure fair comparison in Table \ref{tab:SOTA}, we evaluated our method on the same six categories as other SOTA models. Our approach excels in capturing fine-grained details, achieving top performance on pedestrian crossings, stop lines, car park areas, and dividers, with the second-best result for walkways. While slightly underperforming on large areas like drivable regions, our method (RGC + HSDA) attains the highest mIoU, surpassing all state-of-the-art models by at least 1.6\%.

Among the recent ten state-of-the-art (SOTA) networks, BEVFormer and PETRv2 only present results for two of our six map categories: drivable area and divider. To ensure a fair comparison, we train a variant of the RGC+HSDA network that is restricted to only these classes during training. We also use the results of the single-timestamp BEVFormer model for accurate comparison with our model which does not use temporal information. BEVFormer reports a drivable area accuracy of 80.7\% and a divider accuracy of 21.3\%, which are respectively 0.8\% and 31\% lower than the accuracy of our proposed method. Similar to BEVFormer, PETRv2 also utilizes history frame information and reports 83.3\% accuracy for the drivable area and 44.8\% for dividers. Since PETRv2 does not provide results for a model that does not leverage temporal information, we could only compare our single-frame model with their temporal model. PETRv2 with temporal information achieves 1.8\% higher accuracy on the drivable area but 7.5\% lower accuracy on the divider area compared to our proposed method.

\subsubsection{HSDA for Monocular 3D Object Detection}
\begin{figure}
    \centering
    \includegraphics[width=\linewidth]{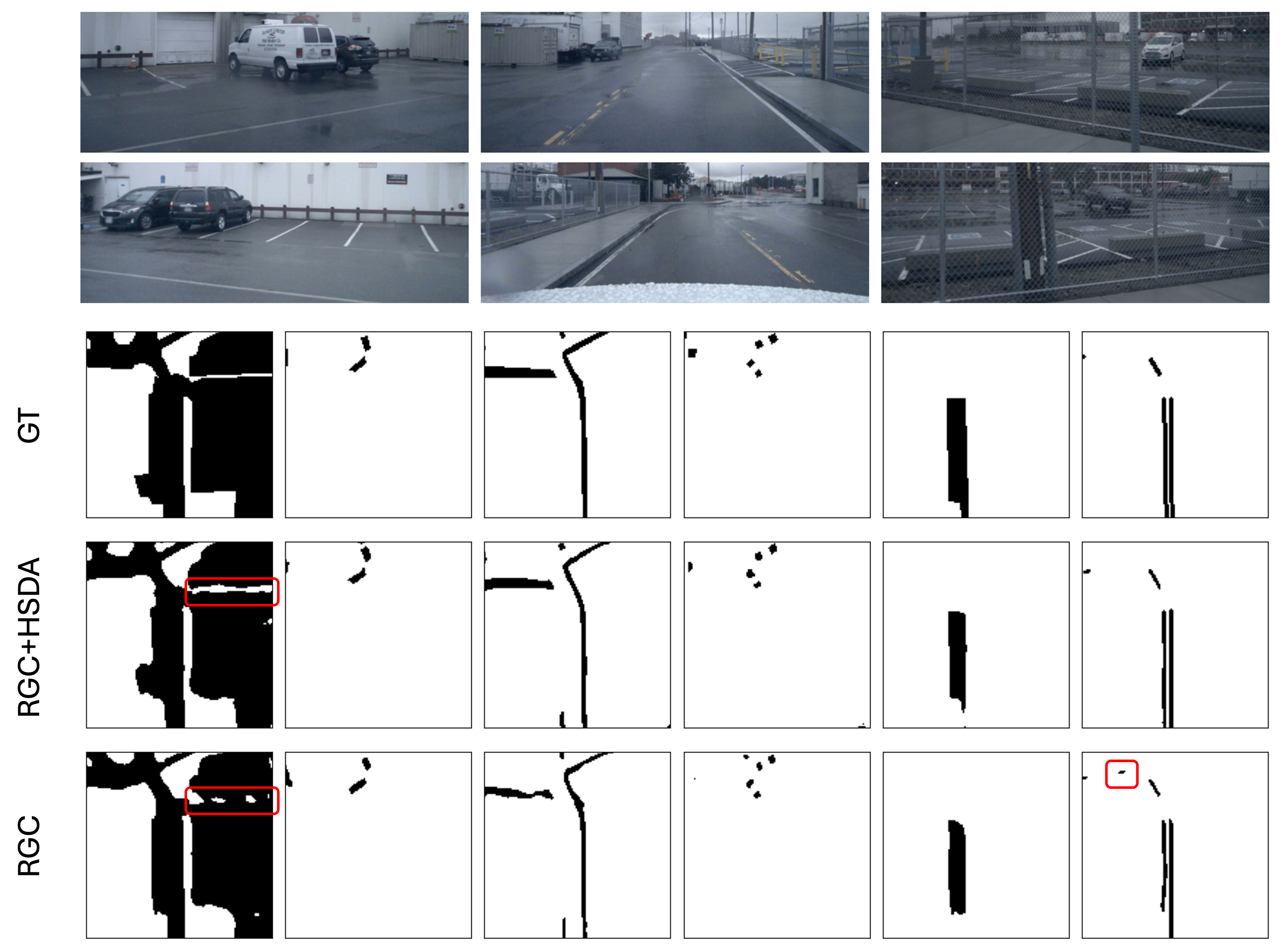}
    \parbox{0.5\textwidth}{\caption{Illustration of one sample rainy scene. Red circles highlight differences between the predicted and ground truth segmentation.}\label{fig:RGC}}
\end{figure}
Though we focus on BEV map segmentation, our method is flexible and can be applied to a wide variety of tasks involving 3D object detection. To explore the applicability of our data augmentation strategy to related fields, we apply HSDA to the monocular 3D object detection task. For this experiment, we use MonoCon\cite{liu2022learning}, a monocular 3D object detection model which exploits aspects of the annotated 3D bounding boxes as auxiliary learning tasks during training.

We conduct the training and evaluation of MonoCon utilizing another popular benchmark, the KITTI dataset\cite{Geiger2012CVPR}.
Our results encompass all three object classes: pedestrian, cyclist, and car, all of which are evaluated on the validation set. For each class, we provide results for the three difficulties of the KITTI benchmark: easy, moderate, and hard. Our metric is the average precision (AP) in 3D space at 40 recall positions. Following the procedure of the KITTI benchmark, we set an IoU threshold of 0.7 for cars, and 0.5 for pedestrians and cyclists.

MonoCon \cite{liu2022learning}, a recent monocular 3D object detector with excellent performance, is chosen as the baseline. 
Subsequently, we retrain the network incorporating the application of HSDA. Table \ref{tab:monocon} compares the baseline MonoCon results and MonoCon with HSDA. The application of HSDA yields a substantial improvement in pedestrian detection, while cyclist detection experiences a minor decline. Car detection also demonstrates a significant improvement. Overall, HSDA contributes to a notable increase in the mean Average Precision (mAP) across all difficulty levels.

\subsection{Qualitative Results}
\begin{figure*}
    \centering
    \includegraphics[width=1\linewidth]{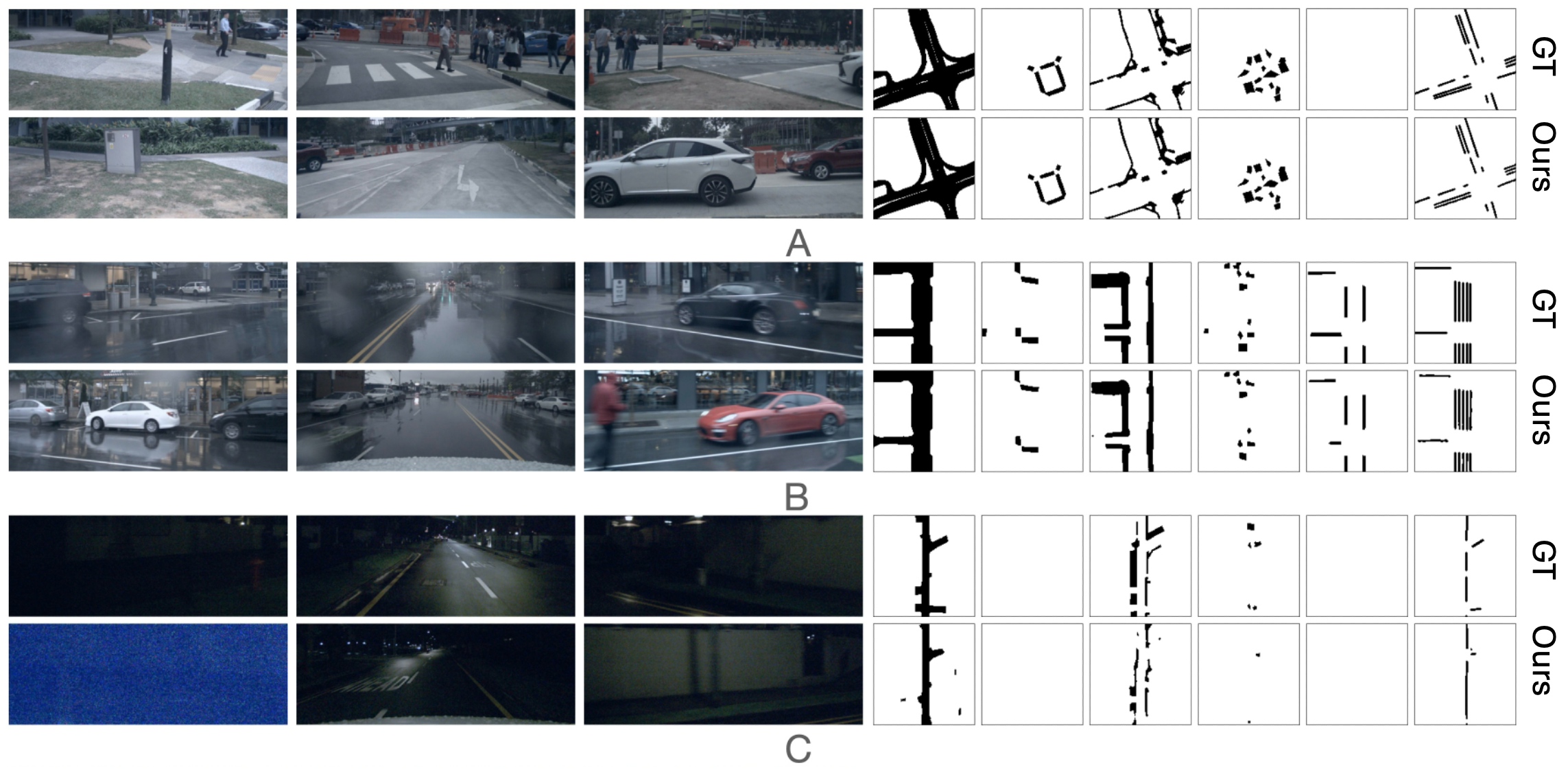}
    \captionsetup{width=1\textwidth}
    \caption{Qualitative results are presented for daytime, rainy, and nighttime scenarios. The left panels display multi-view input images, while the right panels compare ground truth annotations (denoted as "GT") with the output of our proposed method, RGC+HSDA (denoted as "ours"). Six categories are annotated in the right panels: drivable area, pedestrian crossing, walkway, stop line, carpark area, and divider.}
    \label{fig:qualitative}
\end{figure*}

Figure \ref{fig:RGC} illustrates the advantages of the proposed HSDA method based on the RGC model. The first two rows display the camera images of a rainy scene, followed by a comparison of the ground truth with our segmentation results with and without HSDA.
We find that RGC produces false positive segmentation results in both the drivable area and divider, which are resolved by applying HSDA. Notably, even with raindrops on the camera lens, HSDA achieves more accurate results due to its ability to distinguish relevant high-frequency information from noise.

Figure \ref{fig:qualitative} presents input images from three scenes in the nuScenes dataset, along with their corresponding ground truth and segmentation results predicted by our proposed RGC+HSDA model. Scene (A) depicts daytime conditions, (B) rainy conditions, and (C) nighttime conditions. Our model demonstrates satisfactory performance in daytime scenarios, closely aligning with the ground truth. It also performs well in rainy conditions, exhibiting only minor errors at far distances. However, nighttime performance reveals greater uncertainty at far distances, likely due to reduced light. Camera anomalies, such as the unusual blue tint observed in (C), may also contribute to nighttime challenges. Overall, our model yields promising results under daytime and rainy conditions, while we acknowledge room for improvement in nighttime scenarios. 

%-------------------------------------------------------------------------
\section{Conclusion}
\label{sec:conclusion}

This paper investigates the significance of the frequency domain in bird's-eye view (BEV) map segmentation, revealing the particular importance of high-frequency information for network performance. We introduce High-frequency Shuffle Data Augmentation (HSDA), a straightforward but effective method designed to enhance the network's capacity to capture crucial high-frequency information, thereby improving segmentation results for edges and intricate image regions. Our approach is easily implemented and broadly applicable across various models, demonstrating state-of-the-art performance on the nuScenes dataset when applied to RGC. We further demonstrate the applicability of our method to different datasets and perception tasks, namely monocular 3D object detection with KITTI. We anticipate that the findings of this paper will provide valuable insights to the research community and stimulate further exploration of frequency domain applications in autonomous driving.

%-------------------------------------------------------------------------

%%%%%%%%% REFERENCES
{\small
\bibliographystyle{ieee_fullname}
\bibliography{egbib}
}

\end{document}